\documentclass{article}
\pdfoutput=1
\usepackage{times}
\usepackage{graphicx}
\usepackage{subfigure} 
\usepackage{natbib}
\usepackage{algorithm}
\usepackage{algorithmic}
\usepackage{booktabs}
\usepackage{amsmath}
\usepackage{mathrsfs}
\usepackage{amsfonts}
\usepackage{amsthm}
\usepackage{booktabs}
\usepackage{stmaryrd}
\usepackage{url}
\usepackage{relsize}
\usepackage{color}
\usepackage[capitalise]{cleveref}
\usepackage{graphicx}
\def \PG {\textsc{pg}}

\def \SPG {\textsc{spg}}

\def \tr {{\tilde r}}
\def \grad {\nabla}
\def \Deltatheta { {\Delta \theta} }

\newcommand{\cX}{\mathcal{X}}

\newcommand{\cA}{\mathcal{A}}
\newcommand{\cB}{\mathcal{B}}

\newcommand{\cL}{\mathcal{L}}

\newcommand{\Var}{\mathbb{V}}

\newcommand{\intset}[1]{[#1]}

\newcommand{\indic}[1]{\mathbb{I}_{\left [ #1 \right ]}}

\DeclareMathOperator*{\expect}{{\huge \mathbb{E}}}

\def \x {\mathbf{x}}

\def \a {\mathbf{a}}
\def \targ {\mathbf{b}}

\newcommand{\cbar}{\, | \,}
\newcommand{\cdbar}{\, \| \,}

\newcommand{\eqnref}[1]{(\ref{eqn:#1})}

\newcounter{mnote}

\newcommand{\comment}[1]{}

\newcommand{\eg}{e.g.\ }

\newcommand{\flabel}[1]{\label{fig:#1}}
\newcommand{\seclabel}[1]{\label{sec:#1}}

\newcommand{\elabel}[1]{\label{eq:#1}}

\newcommand{\fref}[1]{\cref{fig:#1}}
\newcommand{\sref}[1]{\cref{sec:#1}}

\newcommand{\eref}[1]{\cref{eq:#1}}

\newcommand*\idx[2][]
{ 
\def\next{#1}%
\ifx\empty\next
  (#2)
\else
  (#1, #2)
\fi
}
\newcommand*\elt[3][]
{ 
\def\next{#1}%
\ifx\empty\next
  #2\idx{#3}
\else
  #1\idx{#2,#3}
\fi
}
\newcommand*\pd[3][]
{ 
\def\next{#1}%
\ifx\empty\next
  \frac{\partial#2}{\partial #3}
\else
  \frac{{\partial^{#1} #2}}{\partial#3^{#1}}
\fi
}

\newcommand{\figdir}{figures/}

\newcommand{\fig}[4]
{
\begin{figure}[t!]
\begin{center}
\includegraphics[width=#3\columnwidth]{\figdir#1}
\end{center}
\caption{#4}
\flabel{#2}
\end{figure}
}
\newcommand{\figstar}[4]
{
\begin{figure*}[t!]
\begin{center}
\includegraphics[width=#3\textwidth]{\figdir#1}
\end{center}
\caption{#4}
\flabel{#2}
\end{figure*}
}

\usepackage[accepted]{icml2017}

\icmltitlerunning{Automated Curriculum Learning}

\begin{document} 

\twocolumn[
\icmltitle{Automated Curriculum Learning for Neural Networks}

\icmlauthor{Alex Graves, Marc G. Bellemare, Jacob Menick, R\'emi Munos, Koray Kavukcuoglu\\}{\{gravesa, bellemare, jmenick, munos, korayk\}@google.com}

\icmladdress{Google DeepMind, London UK}

\icmlkeywords{Curriculum learning, instrinsic motivation, neural networks, LSTM}

\vskip 0.3in
]

\begin{abstract}
We introduce a method for automatically selecting the path, or syllabus, that a neural network follows through a curriculum so as to maximise learning efficiency.
A measure of the amount that the network learns from each data sample is provided as a reward signal to a nonstationary multi-armed bandit algorithm, which then determines a stochastic syllabus.
We consider a range of signals derived from two distinct indicators of learning progress: rate of increase in prediction accuracy, and rate of increase in network complexity.
Experimental results for LSTM networks on three curricula demonstrate that our approach can significantly accelerate learning, in some cases halving the time required to attain a satisfactory performance level.
\end{abstract}

\section{Introduction}\seclabel{introduction}
Over two decades ago, in \emph{The importance of starting small}, Elman put forward the idea that a curriculum of progressively harder tasks could significantly accelerate a neural network's training \citep{elman93learning}. However curriculum learning has only recently become prevalent in the field \citep[e.g.,][]{bengio09curriculum}, due in part to the greater complexity of problems now being considered.
In particular, recent work on learning programs with neural networks has relied on curricula to scale up to longer or more complicated programs \cite{sutskever14lte,reed2015neural,graves2016hybrid}.
We expect this trend to continue as the scope of neural networks widens.

One reason for the slow adoption of curriculum learning is that its effectiveness is highly sensitive to the mode of progression through the tasks.
One popular approach is to define a hand-chosen performance threshold for advancement to the next task, along with a fixed probability of returning to earlier tasks, to prevent forgetting \cite{sutskever14lte}.
However, as well as introducing hard-to-tune parameters, this poses problems for curricula where appropriate thresholds may be unknown or variable across tasks.
More fundamentally, it presupposes that the tasks can be ordered by difficulty, when in reality they may vary along multiple axes of difficulty, or have no predefined order at all.

We propose to instead treat the decision about which task to study next as a stochastic policy, continuously adapted to optimise some notion of what \citet{oudeyer07intrinsic} termed \textit{learning progress}.
Doing so brings us into contact with the intrinsic motivation literature \cite{barto13intrinsic}, where various indicators of learning progress have been used as reward signals to encourage exploration, including compression progress \citep{schmidhuber91possibility}, information acquisition \citep{storck1995reinforcement}, Bayesian surprise \citep{itti2009bayesian}, prediction gain \citep{bellemare16unifying} and variational information maximisation \citep{houthooft2016vime}.
We focus on variants of prediction gain, and also introduce a novel class of progress signals which we refer to as complexity gain.
Derived from minimum description length principles, complexity gain equates acquisition of knowledge with an increase in effective information encoded in the network weights.

Given a progress signal that can be evaluated for each training example, we use a multi-armed bandit algorithm to find a stochastic policy over the tasks that maximises overall progress. 
The bandit is nonstationary because the behaviour of the network, and hence the optimal policy, evolves during training.
We take inspiration from a previous work that modelled an adaptive student with a multi-armed bandit in the context of developmental learning \cite{lopes12strategic}. 
Another related area is the field of active learning, where similar gain signals have been used to guide decisions about which data point to label next~\cite{settles2010active}.
Lastly, there are parallels with recent work on using Bayesian optimisation to find the best order in which to train a word embedding network on a language corpus \cite{dyer2016curriculum}; however this differs from our work in that the ordering was entirely determined \textit{before} each training run, rather than adaptively altered in response to the model's progress.

\section{Background}\seclabel{background}
We consider supervised or unsupervised learning problems where target sequences $\targ^1, \targ^2, \dots$ are conditionally modelled given their respective input sequences $\a^1,\a^2, \dots$. For convenience we suppose that the targets are drawn from a finite set $\cB$, noting our framework extends to continuous targets, with densities taking the place of probabilities.
As is typical for neural networks, sequences may be grouped together in batches ($\targ^{1:B}, \a^{1:B})$ to accelerate training. The conditional probability output by the model is
\begin{equation*}
    p(\targ^{1:B} \cbar \a^{1:B})
    = \prod_{i,j} p(\targ_j^i \cbar \targ_{1:j-1}^i, \a_{1:j-1}^i) .
\end{equation*}
From here onwards, we consider each batch as a single example $\x$ from $\cX := (\cA \times \cB)^N$, and write $p(\x):= p(\targ^{1:B} \cbar \a^{1:B})$
for its probability.
Under this notation, a \emph{task} is a distribution $D$ over sequences from $\cX$. 
A \emph{curriculum} is an ensemble of tasks $D_1, \dots, D_N$, and a \emph{sample} is an example drawn from one of the tasks of the curriculum. 
Finally, a \emph{syllabus} is a time-varying sequence of distributions over tasks.

We consider a neural network to be a parametric probabilistic model $p_\theta$ over $\cX$, whose parameters are denoted $\theta$.
The expected loss of the network on the $k^{th}$ task is
\begin{equation*}\elabel{k_loss}
    \cL_k(\theta) := \expect_{\x \sim D_k} L(\x,\theta),
\end{equation*}
where $L(\x,\theta) := -\log p_\theta(\x)$ is the sample loss on $\x$.
Whenever unambiguous, we will simply denote the expected and sample losses by $\cL_k$ and $L(\x)$ respectively.

\subsection{Curriculum Learning}
We consider two related settings. In the \emph{multiple tasks} setting, 
The goal is to perform as well as possible on all tasks in the ensemble $\{ D_k \}$; this is captured by the objective function
\begin{equation*}
    \cL_{\textsc{mt}} := \frac{1}{N}\sum_{k=1}^N \cL_k .
\end{equation*}
In the \emph{target task} setting, we are only interested in minimizing the loss on the final task $D_N$. The other tasks then act as a series of stepping stones to the real problem. The objective function in this setting is simply $\cL_{\textsc{tt}} := \cL_N$.

\subsection{Adversarial Multi-Armed Bandits}\seclabel{bandits}
We view a curriculum containing $N$ tasks as an $N$-armed bandit \citep{bubeck12regret}, and a syllabus as an adaptive policy which seeks to maximize payoffs from this bandit. In the bandit setting, an agent selects a sequence of arms (actions) $a_1 \dots a_T$ over $T$ rounds of play ($a_t \in \{1, \dots, N\}$). After each round, the selected arm yields a payoff $r_t$; the payoffs for the other arms are not observed. 

The classic algorithm for adversarial bandits is Exp3 \citep{auer02nonstochastic},
which uses multiplicative weight updates to guarantee low regret with respect to the best arm. On round $t$, the agent selects an arm stochastically according to a policy $\pi_t$. This policy is defined by a set of weights $w_{t,i}$:
\begin{equation*}
    \pi^{\textsc{exp3}}_t(i) := \frac{e^{w_{t,i}}}{\sum_{j=1}^N  e^{w_{t,j}}} .
\end{equation*}
The weights are the sum of importance-sampled rewards:
\begin{equation*}
    w_{t,i} := \eta \sum_{s < t} \tr_{s,i} \qquad \tr_{s,i} := \frac{r_s \indic{a_s = i}}{\pi_s(i)} .
\end{equation*}
Exp3 acts so as to minimize regret with respect to the single best arm evaluated over the whole history. However, a common occurrence is for an arm to be optimal for a portion of the history, then another arm, and so on; the best strategy is then piecewise stationary. This is generally the case in our setting, as the expected reward for each task changes as the model learns. 
The Fixed Share method \citep{herbster98tracking} addresses this issue by using an $\epsilon$-greedy strategy and mixing in the weights additively. In the bandit setting, this is known as the Exp3.S algorithm (also by \citet{auer02nonstochastic}):
\begin{align}
\pi^{\textsc{exp3.p}}_t(i) &:= (1 - \epsilon) \pi^{\textsc{exp3}}_t(i) + \frac{\epsilon}{N} \label{eqn:fixed_share} \\
w^{\textsc{s}}_{t,i} &:= \log \Big [ (1 - \alpha_t) \exp\left \{w^{\textsc{s}}_{t-1,i} + \eta \tr^\beta_{t-1,i} \right \} \nonumber \\
&+ \frac{\alpha_t}{N-1} \sum_{j \ne i} \exp \left \{w^{\textsc{s}}_{t-1,j} + \eta \tr^\beta_{t-1,j} \right\} \Big ] \quad \nonumber \\
w^{\textsc{s}}_{1,i} &:= 0 \qquad \alpha_t := t^{-1} \qquad     \tr^{\beta}_{s,i} := \frac{r_s \indic{a_s = i} + \beta}{\pi_s(i)} . \nonumber
\end{align}
\subsection{Reward Scaling}\label{sec:reward_scaling}
The appropriate step size $\eta$ depends on the magnitudes of the rewards, which may not be known \emph{a priori}. 
The problem is particularly acute in our setting, where the magnitude depends on how learning progress is measured, and varies over time as the model learns.
To address this issue, we adaptively rescale all rewards to lie in the interval $[-1, 1]$ using the following procedure:
Let $\mathcal{R}_t$ be the history of unscaled rewards up to time $t$, i.e. $\mathcal{R}_t = \{\hat{r}_i\}_{i=1}^{t-1}$. 
Let $q^{\text{lo}}_t$ and $q^{\text{hi}}_t$ be quantiles of $\mathcal{R}_t$, which we choose here to be the 20$^{\text{th}}$ and 80$^{\text{th}}$ percentiles respectively.  
The scaled reward $r_t$ is obtained by clipping $\hat{r}_t$ to the interval $[q^{\text{lo}}_t, q^{\text{hi}_t}]$ and then linearly mapping the result to lie in $[-1, 1]$:
\begin{equation}
    r_t=
\begin{cases}
    -1              & \text{if } \hat{r}_t < q^{\text{lo}}_t \\
    1              & \text{if } \hat{r}_t > q^{\text{hi}}_t \\
    \frac{2 (\hat{r}_t - q^{\text{lo}}_t)}{q^{\text{hi}}_t - q^{\text{lo}}_t} - 1 & \text{otherwise}.\\
\end{cases}
\end{equation}
Rather than keeping the entire history of rewards, we use reservoir sampling to maintain a representative sample, and compute approximate quantiles from this sample. These quantiles can be obtained in $\Theta(log |\mathcal{R}_t|)$ time.

\section{Learning Progress Signals}\seclabel{progress}
Our goal is to use the policy output by Exp3.S as a syllabus for training our models. 
Ideally we would like the policy to maximize the rate at which we minimize the loss, and the reward should reflect this rate -- what \citet{oudeyer07intrinsic} calls \emph{learning progress}.
However, it usually is computationally undesirable or even impossible to measure the effect of a training sample on the target objective, and we therefore turn to surrogate measures of progress. Broadly, these measures are either 1) loss-driven, in the sense that they equate reward with a decrease in some loss; or 2) complexity-driven, when they equate reward with an increase in model complexity.

Training proceeds as follows: at each time $t$, we first sample a task index $k \sim \pi_t$. We then generate a sample from this task, i.e. $\x \sim D_k$. Note that each $\x$ is in general a batch of training sequences, and that in order to reduce noise in the gain signal we draw the whole batch from a single task. 
We compute the chosen measure of learning progress $\nu$ then divide by the time $\tau(\x)$ required to process the sample (since it is the \textit{rate} of progress we are concerned with, and processing time may vary from task to task) to get the raw reward $\hat{r} = \nu/\tau(\x)$
For the purposes of this work, $\tau(\x)$ was simply the length of the longest input sequence in $\x$; for other tasks or architectures a more complex calculation may be required.
We then rescale $\hat{r}$ into a reward $r_t \in [-1, 1]$, and provide it to Exp3.S. The procedure is summarized as Algorithm \ref{algo:syllabus}.

\begin{algorithm}[htb!]
\caption{Intrinsically Motivated Curriculum Learning\label{algo:syllabus}}
\begin{algorithmic}
\medskip
\item[\textbf{Initially:}] $w_i = 0$ for $i \in \intset{N}$
\medskip
\FOR{$t = 1 \dots T$}
    \STATE $\pi(k) := (1 - \epsilon) \frac{e^{w_k}}{\sum_i e^{w_i}} + \frac{\epsilon}{N}$
    \STATE Draw task index $k$ from $\pi$
    \STATE Draw training sample $\x$ from $D_k$
    \STATE Train network $p_\theta$ on $\x$
    \STATE Compute learning progress $\nu$ (Sections \ref{sec:data_progress} \& \ref{sec:model_progress})
    \STATE Map $\hat{r} = \nu/\tau(\x)$ to $r \in [-1, 1]$ (Section \ref{sec:reward_scaling})
    \STATE Update $w_i$ with reward $r$ using Exp3.S \eqnref{fixed_share}
\ENDFOR
\end{algorithmic}
\end{algorithm}

\subsection{Loss-driven Progress}\label{sec:data_progress}
We consider five loss-driven progress signals, all which compare the predictions made by the model before and after training on some sample $\x$. 
The first two signals we present are instantaneous in the sense that they only depend on $\x$. Such signals are appealing because they are typically cheaper to evaluate, and are agnostic about the overall goal of the curriculum. 
The remaining three signals more directly measure the effect of training on the desired objective, but require an additional sample $\x'$. 
In what follows we denote the model parameters before and after training on $\x$ by $\theta$ and $\theta'$ respectively.

\paragraph{Prediction gain (PG).}
Prediction gain is defined as the instantaneous change in loss for a sample $\x$, before and after training on $\x$:
\begin{equation*}
    \nu_{PG} := L(\x, \theta) - L(\x, \theta') .
\end{equation*}
When $p_\theta$ is a Bayesian mixture model, prediction gain upper bounds the model's information gain \citep{bellemare16unifying}, and is therefore closely related to the Bayesian precept that learning is a change in posterior.

\paragraph{Gradient prediction gain (GPG).}
Computing prediction gain requires an additional forward pass. When $p_\theta$ is differentiable, an alternative is to consider the first-order Taylor series approximation to prediction gain:
\begin{equation*}
L(\x, \theta') \approx L(\x, \theta) + \left [ \nabla L(\x, \theta) \right ]^\top \Delta_\theta,
\end{equation*}
where $\Delta_\theta$ is the descent step. Taking this step to be the negative gradient $-\nabla_\theta L(\x, \theta)$ we obtain the gradient prediction gain
\begin{equation*}
    \nu_{GPG} := \| \nabla L(\x, \theta) \|_2^2 .
\end{equation*}
This measures the magnitude of the gradient vector, which has been used an indicator of data salience in the active learning literature \cite{settles2008multiple}.
We will show below that gradient prediction gain is a biased estimate true expected learning progress, and in particular favours tasks whose loss has higher variance.

\paragraph{Self prediction gain (SPG).}
Prediction gain is a biased estimate of the change in $\cL_k(\theta)$, the expected loss on task $k$. Having trained on $\x$, we naturally expect the sample loss $L(\x, \theta)$ to decrease, even though the loss at other points may increase. 
Self prediction gain addresses this issue by sampling a second time from the same task and estimating progress on the new sample:
\begin{equation*}
    \nu_{SPG} := L(\x', \theta) - L(\x', \theta') \qquad \x' \sim D_k .
\end{equation*}

\paragraph{Target prediction gain (TPG).}
We can take the self-prediction gain idea further and evaluate directly on the loss of interest, which has has also been considered in active learning~\cite{roy2001toward}. In the target task setting, this becomes
\begin{equation*}
    \nu_{TPG} := L(\x', \theta) - L(\x', \theta') \qquad \x' \sim D_N .
\end{equation*}
Although this might seem like the most accurate measure so far, it tends to suffer from high variance, and also runs counter to the premise that, early in training, the model cannot improve on the difficult target task and should instead train on a task that it can master.

\paragraph{Mean prediction gain (MPG).}
Mean prediction gain is the analogue of target prediction gain in the multiple tasks setting, where it is natural to evaluate our progress across all tasks. We write
\begin{equation*}
    \nu_{MPG} := L(\x', \theta) - L(\x', \theta') \qquad \; \x' \sim D_k, k \sim U_N,
\end{equation*}
where $U_N$ denotes the uniform distribution over $\{1, \dots, N\}$.
Mean prediction gain has additional variance from sampling an evaluation task $k \sim U_N$.

\subsection{Complexity-driven Progress}\label{sec:model_progress}
So far we have considered gains that gauge the network's learning progress directly, by observing the rate of change in its predictive ability.
We now turn to a novel set of gains that instead measure the rate at which the network's complexity increases.
These gains are underpinned by the Minimum Description Length (MDL) principle \cite{rissanen1986mdl,grunwald2007minimum}: in order to best generalise from a particular dataset, one should minimise the number of bits required to describe the model parameters plus the number of bits required for the model to describe the data.

According to the MDL principle, increasing the model complexity by a certain amount is only worthwhile if it compresses the data by a greater amount.
We would therefore expect the complexity to increase most in response to the training examples from which the network is best able to generalise.
These examples are exactly what we seek when attempting to maximise learning progress.

MDL training for neural networks~\cite{hinton1993keeping} can be practically realised with stochastic variational inference~\cite{graves11vi,kingma2015variational,blundell2015weight}.
In this framework a variational posterior $P_{\phi}(\theta)$ over the network weights is maintained during training, with a single weight sample drawn for each training example.
The parameters $\phi$ of the posterior are optimised, rather than $\theta$ itself.
The total loss is the expected log-loss of the training dataset\footnote{MDL deals with \textit{sets} rather than \textit{distributions}; in this context we consider each $\mathcal{D}_k$ in the curriculum to be a dataset sampled from the task distribution, rather than the distribution itself} (which in our case is the complete curriculum), plus the KL-divergence between the posterior and some fixed~\cite{blundell2015weight} or adaptive~\cite{graves11vi} prior $Q_{\psi}(\theta)$:
\begin{align*}
L_{VI}(\phi,\psi) = \underbrace{KL(P_{\phi} \cdbar Q_{\psi})}_{\text{model complexity}} + \underbrace{\sum\nolimits_k{ \sum_{\x \in D_k}{ \expect_{\theta \sim P_{\phi}} L(\x, \theta)}}}_{\text{data cost}} .
\end{align*}
Since we are using stochastic gradient descent we need to determine the per-sample loss for both the model complexity and the data.
Defining $S := \sum_k |D_k|$ as the total number of samples in the curriculum we obtain
\begin{equation}
\elabel{vi_sample_loss}
L_{VI}(\x, \phi,\psi) := \frac{1}{S} KL(P_{\phi} \cdbar Q_{\psi}) + \expect_{\theta \sim P_{\phi}} L(\x, \theta),
\end{equation}
with $L_{VI}(\phi, \psi) = \sum_k \sum_{\x \sim D_k} L_{VI}(\x, \phi, \psi)$.
Some of the curricula we consider are algorithmically generated, meaning that the number of samples in each task is undefined.
The treatment suggested by the MDL principle is to divide the complexity cost by the total number of samples generated so far.
However we simplified matters by setting $S$ to a large constant that roughly matches the number of samples we expect to see during training.

We used a diagonal Gaussian for both P and Q, allowing us to determine the complexity cost analytically:
\begin{equation*}
KL(P_\phi\cdbar Q_\psi) = \frac{(\mu_\phi - \mu_\psi)^2 + \sigma_\phi^2 - \sigma_\psi^2}{2 \sigma_\psi^2} + \ln\left(\frac{\sigma_\psi}{\sigma_\phi}\right),
\end{equation*}
where $\mu_\phi, \sigma^2_\phi$ and $\mu_\psi, \sigma^2_\psi$ are the mean and variance vectors for $P_{\phi}$ and $Q_{\psi}$ respectively.
We adapted $\psi$ with gradient descent along with $\phi$, and
the gradient of $\expect_{\theta \sim P_{\phi}} L(\x, \theta)$ with respect to $\phi$ was estimated using the reparameterisation trick\footnote{The reparameterisation trick yields a better gradient estimator for the posterior variance than that used in~\cite{graves11vi}, which requires either calculation of the diagonal of the Hessian, or a biased approximation using the empirical Fisher. The gradient estimator for the posterior mean is the same in both cases.}~\cite{kingma14vae} with a single Monte-Carlo sample. 
The SoftPlus function ${y=\ln(1+e^{x})}$ was used to ensure that the variances were positive~\cite{blundell2015weight}.
\paragraph{Variational complexity gain (VCG).}
The increase of model complexity induced by a training example can be estimated from the change in complexity following a single parameter update from $\phi$ to $\phi'$ and $\psi$ to $\psi'$, yielding
\begin{equation*}
\nu_{VCG} := KL(P_{\phi'} \cdbar Q_{\psi'}) - KL(P_{\phi} \cdbar Q_{\psi})
\end{equation*}
\paragraph{Gradient variational complexity gain (GVCG).}
As with prediction gain, we can derive a first order Taylor approximation using the direction of gradient descent:
{\small
\begin{align*}
&KL(P_{\phi'} \cdbar Q_{\psi'}) \approx KL(P_{\phi} \cdbar Q_{\psi}) \\&\qquad- \left [ \nabla_{\phi,\psi} KL(P_{\phi} \cdbar Q_{\psi}) \right ]^\top  \nabla_{\psi,\phi}\cL_{MDL}(\x, \phi,\psi)\\
&\implies \nu_{VCG} \approx C - \left[\nabla_{\phi,\psi} KL(P_{\phi} \cdbar Q_{\psi})\right]^\top \nabla_{\phi} \expect_{\theta \sim P_{\phi}} L(\x, \theta),
\end{align*}
}
where $C$ is a term that does not depend on $\x$ and is therefore irrelevant to the gain signal.
We define the gradient variational complexity gain as
\begin{equation*}
\nu_{GVCG} := \left[\nabla_{\phi,\psi} KL(P_{\phi} \cdbar Q_{\psi})\right]^\top \nabla_{\phi} \expect_{\theta \sim P_{\phi}} L(\x, \theta),
\end{equation*}
which is the directional derivative of the $KL$ along the gradient descent direction. 
We believe that the linear approximation is more reliable here than for prediction gain, as the model complexity has less curvature than the loss surface.

\paragraph{Relationship to VIME.}
Variational Information Maximizing Exploration (VIME) ~\cite{houthooft2016vime}, uses a reward signal that is closely related to variational complexity gain.
The difference is that while VIME measures the $KL$ between the posterior before and after a step in parameter space, we consider the change in KL between the posterior and prior induced by the step. 
Therefore, while VIME looks for any change to the posterior, we focus only on changes that alter the divergence from the prior.
Further research will be needed to assess the relative merits of the two signals.

\paragraph{L2 gain (L2G).}
Variational inference tends to slow down learning, making it appealing to define a complexity-based progress signal applicable to more conventionally trained networks.
Many of the standard neural network regularisation terms, such as Lp-norms, can be viewed as defining an upper bound on model description length~\cite{graves11vi}.
We therefore hypothesize that the increase in regularisation cost will be indicative of the increase in model complexity.
To test this hypothesis we consider training with a standard L2 regularisation term added to the loss:
\begin{equation}
\elabel{l2_loss}
L_{L2}(\x, \theta) = L(\x, \theta) + \frac{\alpha}{2} \| \theta \|_2^2
\end{equation}
where $\alpha$ is an empirically chosen constant. In this case the complexity gain can be defined as
\begin{equation}
\nu_{L2G} :=  \| \theta' \|_2^2 - \| \theta \|_2^2
\end{equation}
where we have dropped the $\alpha/2$ term as the gain will anyway be rescaled to $[-1,1]$ before use.
The corresponding first-order approximation is
\begin{equation}
\nu_{GL2G} := \left[\theta\right]^\top  \nabla_{\theta} L(\x, \theta)
\end{equation}
It is possible to calculate L2 gain for unregularized networks; however we found this an unreliable signal, presumably because the network has no incentive to decrease complexity when faced with uninformative data.

\subsection{Prediction Gain Bias}
\seclabel{bias}
Prediction gain, self prediction gain and gradient prediction gain are all closely related, but incur varying degrees of bias and variance.
We now present a formal analysis of the biases present in each, noting that a similar treatment can be applied to our complexity gains.

We assume that the loss $L$ is locally well-approximated by its first-order Taylor expansion:
\begin{equation}
    L(\x, \theta') \approx L(\x, \theta) + \grad L(\x, \theta)^\top \Delta \theta \label{eqn:taylor_approximation}
\end{equation}
where $\Delta \theta := \theta' - \theta$.
For ease of exposition, we also suppose the network is trained with stochastic gradient descent (the same argument leads to similar conclusions when consider higher-order optimization methods):
\begin{equation}
    \Deltatheta := -\alpha \grad L(\x, \theta). \label{eqn:descent_step}
\end{equation}
We define the true expected learning progress as
\begin{equation*}
    \nu := \expect_{\x' \sim D} \left [ \cL(\theta) - \cL(\theta') \right ] = \alpha \big \| \expect_{\x' \sim D} \grad L(\x, \theta)  \big\|^2,
\end{equation*}
with the identity following from \eqnref{descent_step} (recall that $\cL(\theta) = \expect_\x L(\theta)$).
The expected prediction gain is then
\begin{equation*}
    \nu_\PG = \expect_{\x' \sim D} \left [ L(\x, \theta) - L(\x, \theta') \right ] = \alpha \expect_{\x' \sim D}  \big\| \grad L(\x, \theta)  \big \|^2 .
\end{equation*}
Defining
\begin{equation*}
    \Var \big (\grad L(\x, \theta) \big) := \expect \big \| \grad L(\x, \theta) - \expect \grad L(\x', \theta) \|^2,
\end{equation*}
we find that prediction gain is the sum of two terms: true expected learning progress, plus the gradient variance:
\begin{equation*}
    \nu_\PG = \nu + \Var \big (\grad L(\x, \theta) \big) .
\end{equation*}
We conclude that \emph{for equal learning progress, a prediction gain-based curriculum maximizes variance}.
The problem is made worse when using gradient prediction gain, which actually relies on the Taylor approximation \eqnref{taylor_approximation}.
On the other hand, self prediction gain is an unbiased estimate of expected learning progress:
\begin{equation*}
    \expect_{\x} \nu_\SPG = \expect_{\x, \x' \sim D} \left [ L(\x', \theta) - L(\x', \theta') \right ] = \nu .
\end{equation*}
Naturally, its use of two samples results in higher variance than prediction gain, suggesting a bias-variance trade off between the two estimates.

\section{Experiments}\seclabel{experiments}
To test the efficacy of our approach, we applied all the gains defined in the previous section to three task suites: synthetic language modelling on text generated by n-gram models, repeat copy~\cite{graves14ntm} and the bAbI tasks~\cite{weston15babi}

The network architecture was stacked unidirectional LSTM~\cite{graves2013generating} for all experiments, and the training loss was cross-entropy with either categorical targets and softmax output, or Bernoulli targets and sigmoid outputs, optimised by RMSProp with momentum \cite{hinton12rmsprop,graves2013generating}, using a momentum of 0.9 and a learning rate of $10^{-5}$ unless specified otherwise.
The parameters for the Exp3.S algorithm were $\eta = 10^{-3}$, $\beta = 0$, $\epsilon=0.05$.
For all experiments, one set of networks was trained with variational inference (VI) to test the variational complexity gain signals, and another set was trained with normal maximum likelihood (ML) for the other signals; both sets were repeated 10 times with different random seeds to initialise the network weights.
The $\alpha$ regularisation parameter from \eref{l2_loss} for the networks trained with L2 gain signals was $10^{-4}$ for all experiments. 
For all plots with a time axis, time is defined as the total number of input steps processed so far.
In the absence of hand-designed curricula for these tasks, our performance benchmarks are 1) a fixed uniform policy over all the tasks and 2) directly training on the target task (where applicable).
All losses and error rates are measured on independent samples not used for training or reward calculation. 

\subsection{N-Gram Language Modelling}
Our first experiment aims to illustrate and compare the behaviour induced by different gains. We trained character-level Kneser-Ney n-gram models \citep{KneserNey95} on the King James Bible data from the Canterbury corpus \citep{arnold1997corpus},
with the maximum depth parameter $n$ ranging between $0$ to $10$. We then used each model to generate a separate dataset of 1M characters, which we divided into 
disjoint sequences of 150 characters. The first 50 characters of each sequence were used as burn-in context for the next 100, which the network was trained to predict.
The LSTM network had two layers of 512 cells, and the batch size was 32.

An important characteristic of this dataset is that the amount of linguistic structure increases monotonically with $n$. Simultaneously, the entropy -- and hence, minimum achievable loss -- decreases almost monotonically in $n$. 
If we believe that learning progress should be higher for interesting data than for data that is difficult to predict, we would expect the gain signals to be drawn to higher $n$: they should favour structure over noise.
We note that in this experiment the curriculum is superfluous: the most efficient strategy for learning the 10-gram source is to directly train on it.

\fig{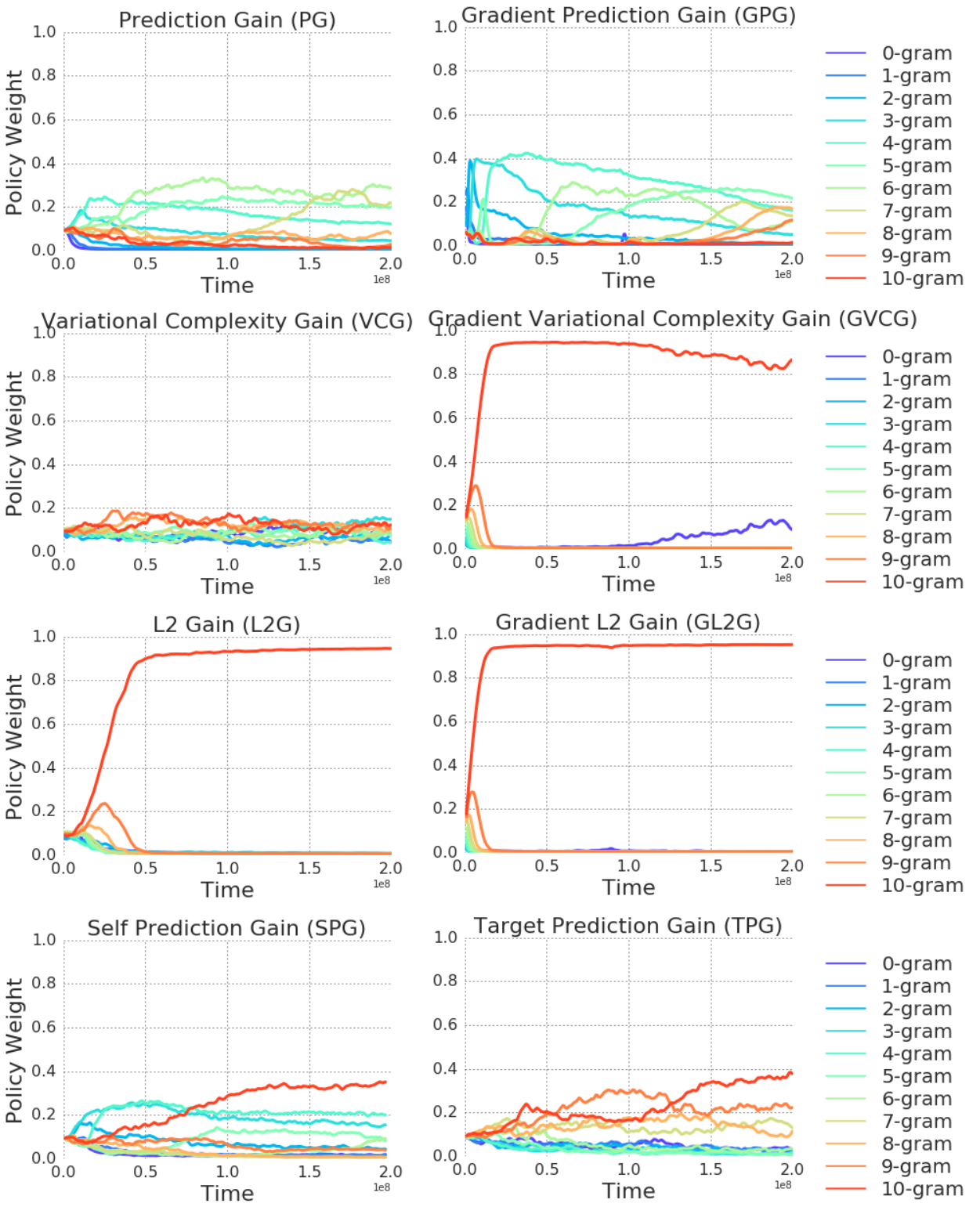}{ngram_policy}{1}{N-gram policies for different gain signals, truncated at $2\times10^8$ steps. All curves are averages over 10 runs}

\fref{ngram_policy} shows that most of the complexity-based gain signals from \sref{model_progress} (L2G, GL2G, GVCG) progress rapidly through the curriculum before focusing strongly on the 10-gram task (though interestingly, GVCG appears to revisit 0-gram later on in training). 
The clarity of the result is striking, given that sequences generated from models beyond about 6-gram are difficult to distinguish by eye. 
VCG follows a similar path, but with much less confidence, presumably due to the increased noise.
The loss-based measures (PG, GPG, SPG, TG) also tend to move towards higher n, although more slowly and with less certainty. 
Unlike the complexity gains, they tend to initially favour the lower-n tasks, which may be desirable as we would expect early learning to be more efficient with simpler data.

\subsection{Repeat Copy}

In the repeat copy task~\cite{graves14ntm} the network receives an input sequence of random bit vectors, and is then asked to output that sequence a given number of times.
The task has two main dimensions of difficulty: the length of the input sequence and the required number of repeats, both of which increase the demand on the models memory.
Neural Turing machines are able to learn a `for-loop' like algorithm on simple examples that can directly generalise to much harder examples~\cite{graves14ntm}.
For LSTM networks without access to external memory, however, significant retraining is required to adapt to harder tasks.

We devised a curriculum with both the sequence length and the number of repeats varying from 1 to 13, giving 169 tasks in all, with length 13, repeats 13 defined as the target task.
The LSTM network had a single layer of 512 cells, and the batch size was 32.
As the data was generated online, the number of samples $S$ in \eref{vi_sample_loss} (the per-sample VI loss) was undefined; we arbitrarily set it to 169M (1M per task in the curriculum).

\fig{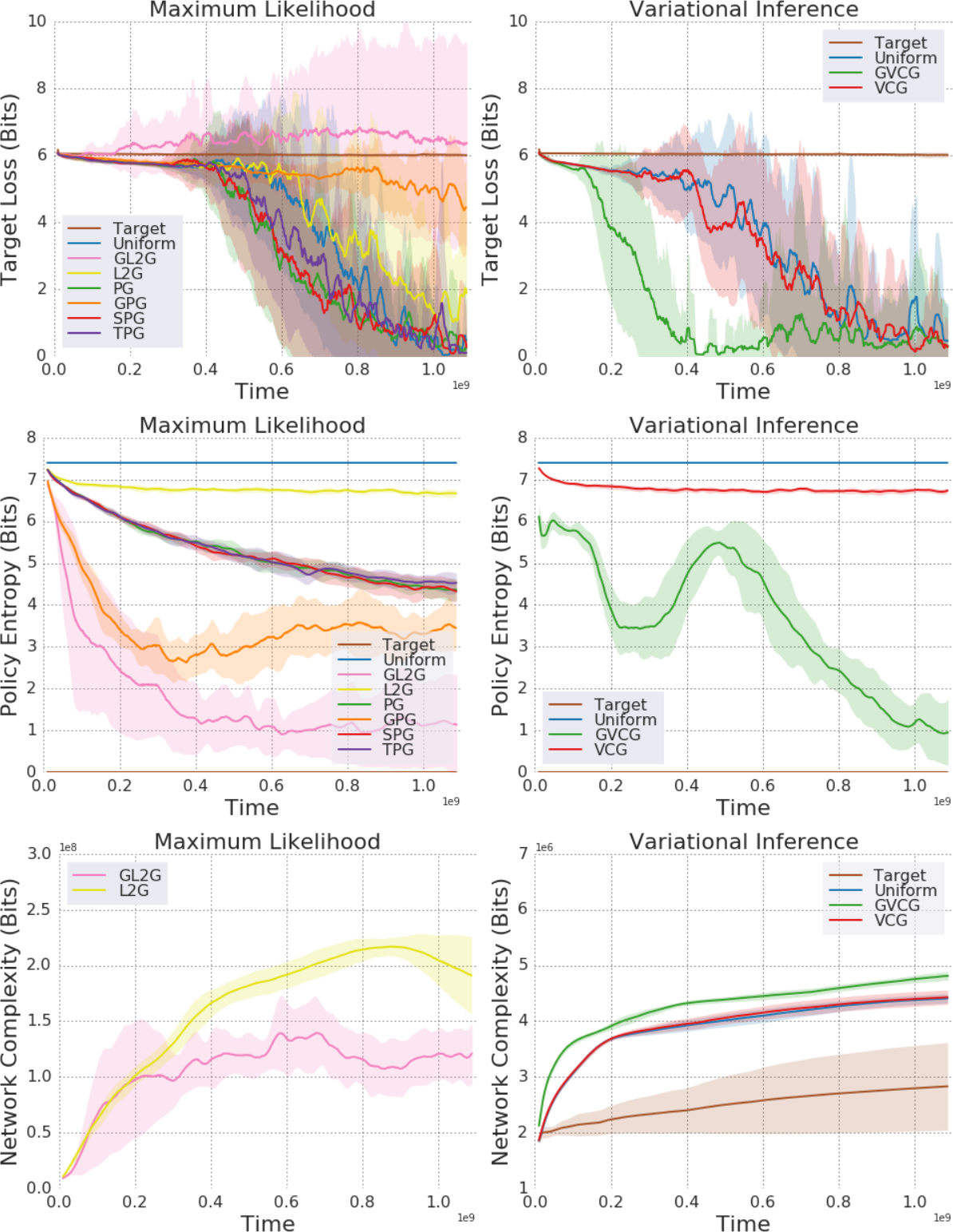}{repeat_copy_performance}{1}{Target task loss (per output), policy entropy and network complexity for the repeat copy task, truncated at $1.1 \times 10^9$ steps. Curves are averages over 10 runs, shaded areas show the standard deviation. Network complexity was computed by multiplying the per-sample complexity cost by the total size of the training set.}

\fref{repeat_copy_performance} shows that GVCG solves the target task about twice as fast as uniform sampling for VI training, and that the PG, SPG and TPG gains are somewhat faster than uniform for ML training, especially in the early stages.
From the entropy plots it is clear that these signals all lead to strongly non-uniform policies.
The VI complexity curves also demonstrate that GVCG yields significantly higher network complexity than uniform sampling,
supporting our hypothesis that increased complexity correlates with learning progress. 
Unlike GVCG, the VCG signal did not deviate far from a uniform policy,.
L2G and particularly GPG and GL2G were much worse than uniform, suggesting that (1) the bias induced by the gradient approximation has a pernicious effect on learning and (2) that the increase in L2 norm is not a reliable measure of increased network complexity.
Training directly on the target task failed to learn at all, underlining the necessity of curriculum learning for this problem.

\figstar{loop_policy_loss}{loop_policy}{0.95}{Average policy and loss per output over time for GVCG networks on the repeat copy task. Plots were made by dividing the first $4\times10^8$ steps into five equal bins, then averaging over the policies of all 10 networks over all times within each bin.}

\fref{loop_policy} reveals a consistent strategy for the GVCG syllabuses, first focusing on short sequences with high repeats, then long sequences with low repeats, thereby decoupling the two dimensions of difficulty.
At each stage the loss is substantially reduced across many tasks that the policy does not focus on.
Crucially, this means that the network does not have to visit each of the 169 tasks to solve them all, and the syllabus is able to exploit this fact to more efficiently complete the curriculum.

\subsection{Babi}
The bAbI dataset~\cite{weston15babi} consists of 20 synthetic question-answering problems designed to probe the basic reasoning capabilities of machine learning models.
Although bAbI was not specifically designed for curriculum learning,
some of the tasks follow a natural ordering of complexity (\eg `Two Arg Relations', `Three Arg Relations') and all are based on a consistent probabilistic grammar, leading us to hope that an efficient syllabus could be found for learning the whole set.
The usual performance measure for bAbI is the number of tasks `completed' by the model, where completion is defined as getting less than 5\% of the test set questions wrong. 

The data representation followed~\cite{graves2016hybrid}, with each word in the observation and target sequences represented as a 1-hot vector, along with an extra binary channel to mark answer prompts.
The original datasets were small, with either 1K or 10K questions per task, so as to test generalisation from limited samples.
However LSTM is known to perform poorly in this setting~\cite{sukhbaatar2015end,graves2016hybrid}, and we wished to avoid the confounding effect of overfitting on curriculum learning.
We therefore used the bAbI code~\cite{weston15babi} to generate 1M stories (each containing one or more questions) for each of the 20 tasks.
With so many examples, we found that training and evaluation set performance were indistinguishable, and therefore report training performance only.
The LSTM network had two layer of 512 cells, the batch size was 16, and the RMSProp learning rate was $3\times 10^{-5}$.

\fig{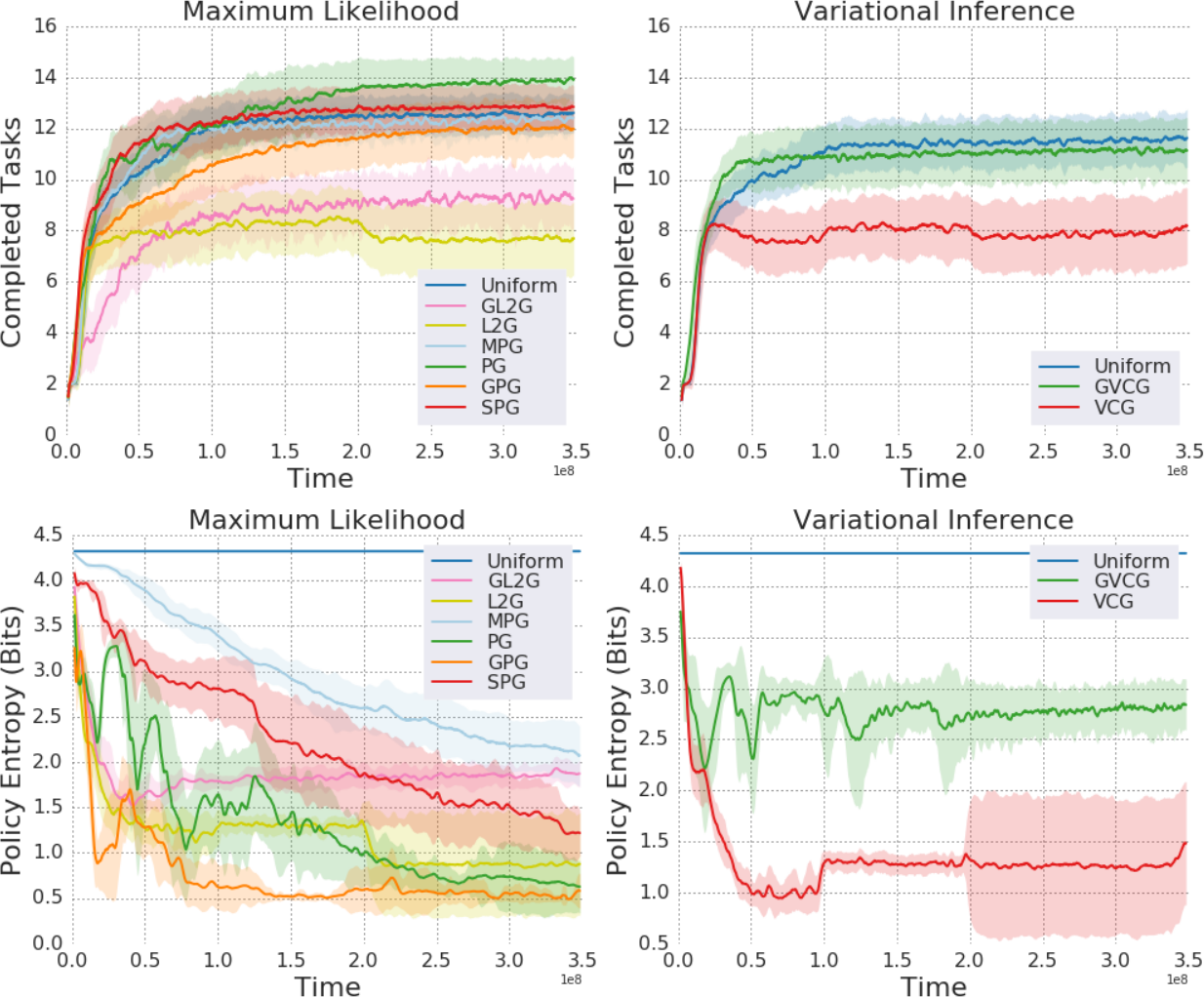}{babi_performance}{1}{Completion and entropy curves for the bAbI curriculum, truncated at $3.5\times10^8$ steps. Curves are means over ten runs, shaded areas show standard deviation.}

\fref{babi_performance} shows that prediction gain (PG) clearly improved on uniform sampling in terms of both learning speed and number of tasks completed; for self-prediction gain (SPG) the same benefits were visible, though less pronounced.
The other gains were either roughly equal to or worse than uniform.
For variational inference training, GVCG was faster than uniform at first, then slightly worse later on, while VCG performed  poorly for reasons that are unclear to us.
In general, training with variational inference appeared to hamper progress on the bAbI tasks.

\fig{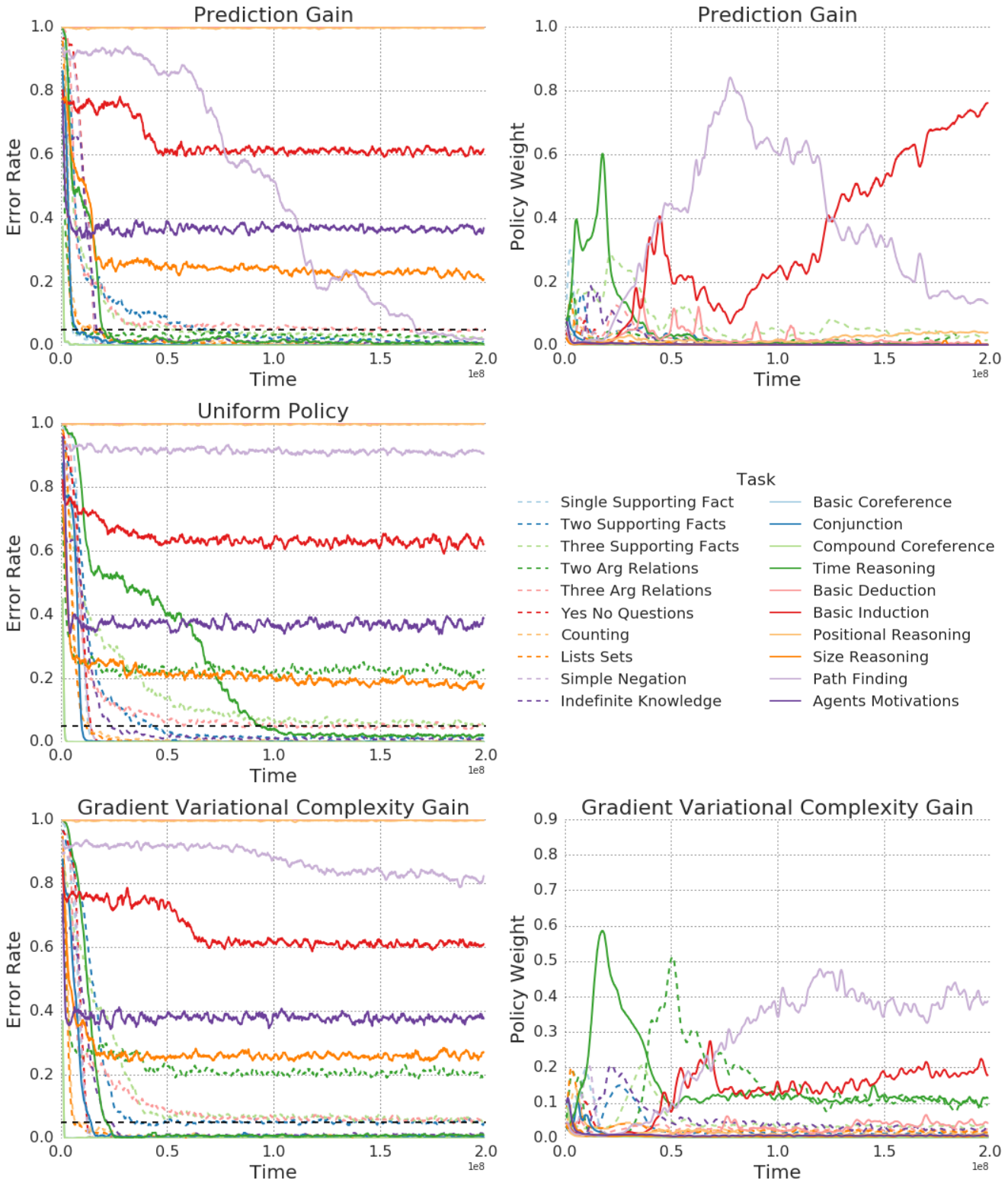}{babi_learning}{1}{Per-task policy and error curves for bAbI, truncated at $2\times10^8$ steps. All plots are averaged over 10 runs. Black dashed lines show the 5\% error threshold for task completion.}

\fref{babi_learning} shows how the PG and GVCG syllabuses accelerate the network's progress by selectively focusing on specific tasks until completion.
For example, they both solve `Time Reasoning' much faster than uniform sampling by concentrating on it early in training; similarly, PG focuses strongly on `Path Finding' (one of the harder bAbI tasks) until it solves it.
Also noteworthy is the way the syllabuses progress from `Single Supporting Fact' to `Three Supporting Facts' in order; this shows that our gain signals can discover implicit orderings, and hence opportunities for efficient transfer, in an unsorted curriculum.

\section{Conclusion}\seclabel{conclusion}
Our experiments suggest that using a stochastic syllabus to maximise learning progress can lead to significant gains in curriculum learning efficiency, so long as a suitable progress signal is used. 
We note however that uniformly sampling from all tasks is a surprisingly strong benchmark. 
We speculate that this is because learning is dominated by gradients from the tasks on which the network is making fastest progress, inducing a kind of implicit curriculum, albeit with the inefficiency of unnecessary samples.
For maximum likelihood training, we found prediction gain to be the most consistent signal, while for variational inference training, gradient variational complexity gain performed best. 
Importantly, both are instantaneous, in the sense that they can be evaluated using only the samples used for training.
As well as being more efficient, this has broader applicability to tasks where external evaluation is difficult, and suggests that learning progress is best assessed on a local, rather than global basis.

\bibliographystyle{apalike}
\bibliography{curriculum}

\end{document}